\documentclass[letterpaper]{article} 
\usepackage{aaai23}  
\usepackage{times}  
\usepackage{helvet}  
\usepackage{courier}  
\usepackage[hyphens]{url}  
\usepackage{graphicx} 
\usepackage{natbib}  
\usepackage{caption} 
\usepackage{algorithm}
\usepackage{algorithmic}

\usepackage{newfloat}
\usepackage{listings}
\DeclareCaptionStyle{ruled}{labelfont=normalfont,labelsep=colon,strut=off} 
\floatstyle{ruled}
\newfloat{listing}{tb}{lst}{}
\floatname{listing}{Listing}
\usepackage{amssymb}
\usepackage{amsmath}
\usepackage[title]{appendix}
\usepackage{bbding}
\usepackage{makecell}
\usepackage{multirow}

\DeclareMathOperator*{\argmin}{arg\,min}

\title{Multi-Scale Control Signal-Aware Transformer for\\ Motion Synthesis without Phase}
\author{
    Lintao Wang\textsuperscript{\rm 1}, Kun Hu\textsuperscript{\rm 1,}\thanks{Corresponding author.}, Lei Bai\textsuperscript{\rm 2}, Yu Ding\textsuperscript{\rm 3}, Wanli Ouyang\textsuperscript{\rm 1}, Zhiyong Wang\textsuperscript{\rm 1}
}
\affiliations{
    \textsuperscript{\rm 1}The University of Sydney,
    \textsuperscript{\rm 2}Shanghai AI Laboratory,
    \textsuperscript{\rm 3}Netease Fuxi AI Lab\\
    lwan3720@uni.sydney.edu.au, kun.hu@sydney.edu.au, baisanshi@gmail.com, dingyu01@corp.netease.com, wanli.ouyang@sydney.edu.au, zhiyong.wang@sydney.edu.au
}

\begin{document}
\maketitle
\begin{abstract}
Synthesizing controllable motion for a character using deep learning has been a promising approach due to its potential to learn a compact model without laborious feature engineering. To produce dynamic motion from weak control signals such as desired paths, existing methods often require auxiliary information such as phases for alleviating motion ambiguity, which limits their generalisation capability. As past poses often contain useful auxiliary hints, in this paper, we propose a task-agnostic deep learning method, namely Multi-scale Control Signal-aware Transformer (MCS-T), with an attention based encoder-decoder architecture to discover the auxiliary information implicitly for synthesizing controllable motion without explicitly requiring auxiliary information such as phase. 
Specifically, an encoder is devised to adaptively formulate the motion patterns of a character's past poses with multi-scale skeletons, and  a decoder driven by control signals to further synthesize and predict the character's state by paying context-specialised attention to the encoded past motion patterns. 
As a result, it helps alleviate the issues of low responsiveness and slow transition which often happen in conventional methods not using auxiliary information.
Both qualitative and quantitative experimental results on an existing biped locomotion dataset, which involves diverse types of motion transitions, demonstrate the effectiveness of our method. In particular, MCS-T is able to successfully generate motions comparable to those generated by the methods using auxiliary information.
\end{abstract}

\section{Introduction}
Interactively controlling a character has been increasingly demanded by various applications such as gaming, virtual reality and robotics. This task remains challenging to achieve realistic and natural poses with complex motions and environments, even with large amount of high quality motion capture (MoCap) data for modelling \cite{PFNN,deepmimic}. 
Recently, deep learning techniques have been studied for controllable motion synthesis given their strong learning capability yet efficient parallel structures for fast runtime. Many encouraging results have been achieved using deep architectures such as multilayer perceptron (MLP) networks \cite{PFNN}, recurrent neural networks \cite{controllableRNN}, generative networks \cite{moglow} and deep reinforcement learning architectures \cite{deepmimic}. 
Particularly, due to the potentials of delivering fast, responsive yet high-quality controllers, MLP networks have been devised for biped locomotion \cite{PFNN}, quadruped locomotion \cite{MANN}, daily interaction \cite{NSM}, basketball play \cite{localphase} and stylised motion prediction \cite{stylemodel}. 
Since a weak control signal, which is commonly used in graphics, often corresponds to a large variation of possible motions, these studies have to rely on auxiliary phase variables in line with the character's contact states for disambiguation purposes. However, the contact states may not be available for all kinds of motions and may require manual correction during data acquisition. 
By contrast, recurrent neural networks, e.g.~\cite{controllableRNN}, aim to constrain the next pose prediction subject to the past motions, which can be task-agnostic in terms of motion category and demonstrate better generalisation capability. The key limitation of RNN based methods is that they often suffer from slow responsiveness issues due to the large variation of the hidden memory \cite{NSM}. 

\begin{figure}[t]
    \centering
    \includegraphics[width=\linewidth]{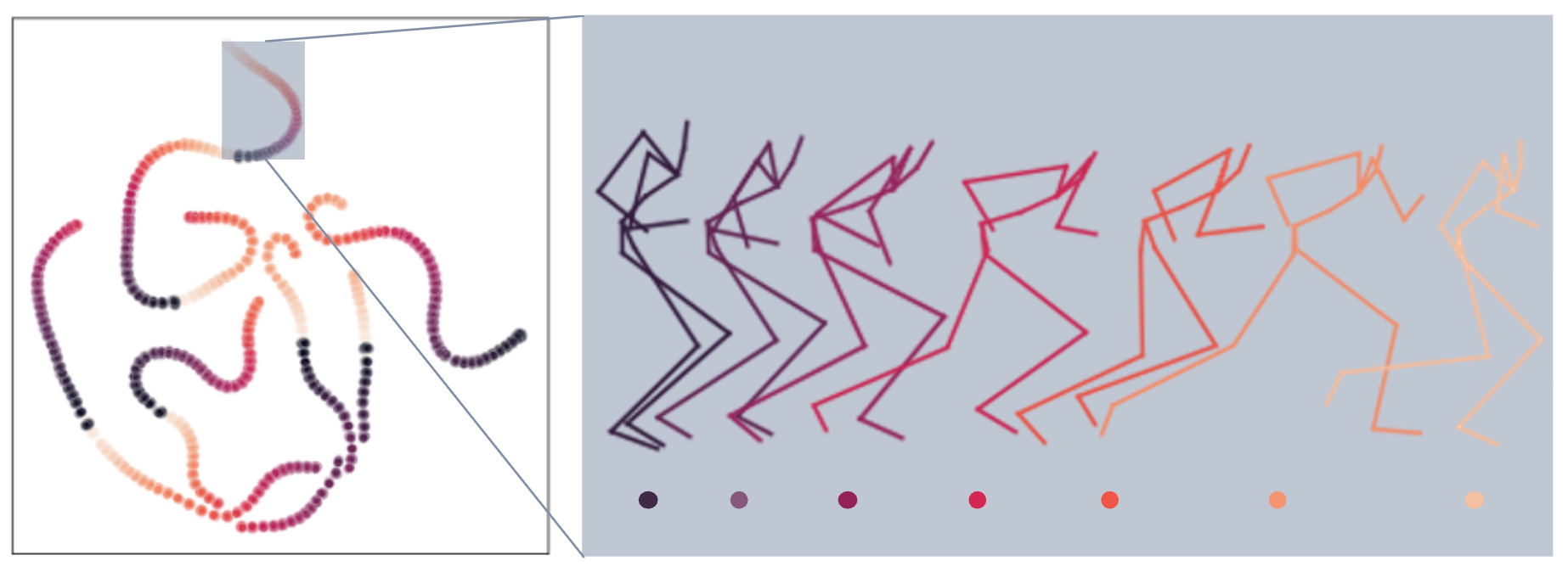}
    \caption{Illustration of the motion manifold of a running motion sequence. Joint positions and velocities of individual poses are projected into a 2D space by t-SNE and colored in line with their phases. It is noticed that auxiliary phases are continuously distributed on the manifold, which suggests the potential of inferring the phases from motion attributes.}
    \label{fig:latent}
\end{figure}

We believe that auxiliary information can be inferred from a character's past motions. As shown in Figure \ref{fig:latent}, a walking motion sequence is represented in 2D manifolds by using different attributes (e.g. joint positions and velocities). It can be observed that the phases are continuously distributed on the manifolds, which enables the auxiliary information inference from the motion related attributes. 
Nonetheless, the past poses should be used ``attentionally" since not all of them are always informative especially during motion transition, which is the reason that RNN based methods perform poorly without explicit data augmentation to the transitional cases in motion capture (MoCap) data. Therefore, in this work, we aims to study a deep learning based task-agnostic method to produce dynamic motion from trajectory-based control signals, without explicitly using additional auxiliary information such as phase.

Specifically, we propose a transformer-based encoder-decoder architecture, namely Multi-Scale Control Signal-aware Transformer (MCS-T), to attend to the motion information of past poses and trajectory with respect to various scenarios.
An encoder formulates the past motion patterns of a character from multi-scale skeletons in pursuit of learning spatio-temporal patterns from different levels of dynamics. With the past motion information, the encoder is expected to formulate conventional auxiliaries implicitly. Then, a decoder guided by the control signals synthesizes and predicts the next character pose by paying trajectory-specialised attention to the encoded historical motion patterns, rather than using a long inflexible memory setting. This dynamic motion modelling pipeline helps alleviate the issues of low responsiveness and slow transition, which can be observed in existing methods not using auxiliary information. 
Comprehensive experiments on a biped locomotion dataset containing various motion transitions (e.g., sudden jumping and uneven terrain walking) demonstrate the effectiveness of MCS-T. It produces responsive and dynamic motion, and achieves a performance comparable to that of the methods explicitly using auxiliary information while retaining such capability for various motion categories. 

The main contributions of this paper can be summarised as follows: 
\begin{itemize}
    \item A novel real-time motion controller, namely Multi-Scale Control Signal-aware Transformer, is proposed to improve the responsiveness and motion dynamics over existing methods not explicitly using auxiliary information. 
    It is also task-agnostic, compared with the methods explicitly using auxiliary information. 
    To the best of our knowledge, our task-agnostic method is one of the first studies utilising transformer based encoder-decoder scheme for controllable motion synthesis.
    \item A multi-scale graph modelling scheme is devised to exploit rich skeleton dynamics.
    \item A novel control signal-aware self-attention is devised to inject control signals for motion prediction.
    \item Comprehensive experiments were conducted to demonstrate the effectiveness of our proposed MCS-T. 
\end{itemize}

\section{Related Work}

In this section, we review related studies in terms of kinematic-based controllable motion synthesis, transformer based motion learning and multi-scale skeleton. 

\begin{figure*}[ht]
    \centering
    \includegraphics[width=\textwidth]{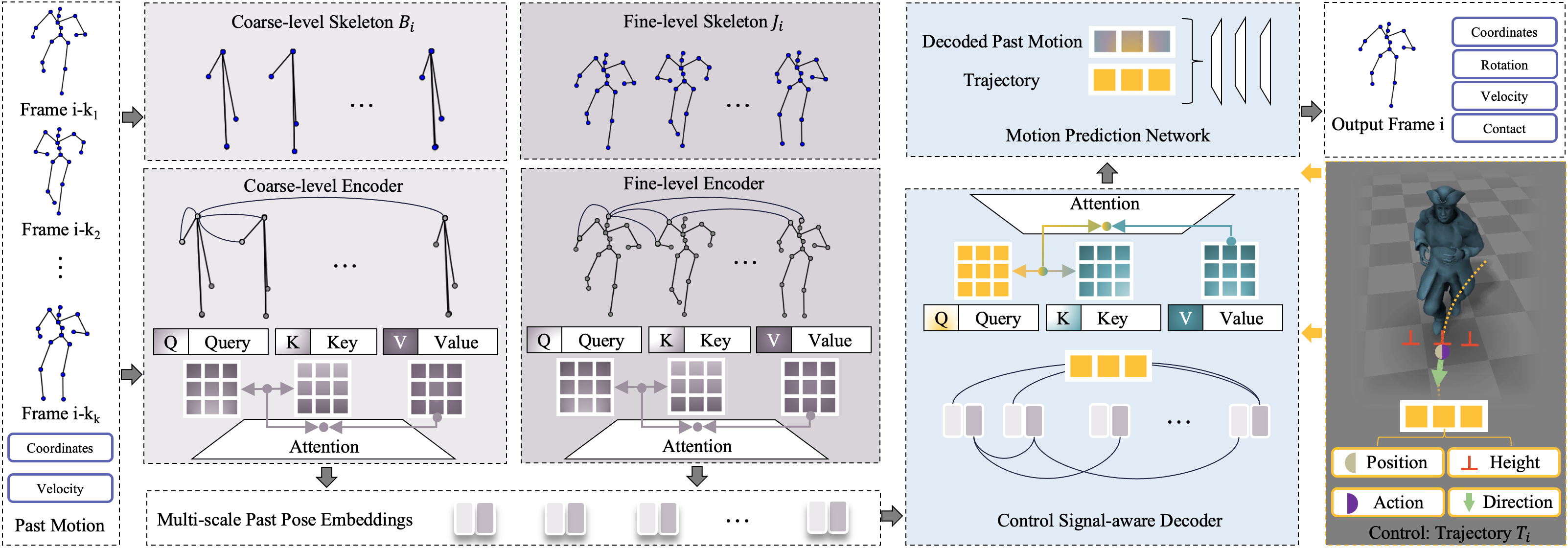}
    \caption{Illustration of our proposed MCS-T method, which is based on an encoder-decoder architecture to formulate the past motion patterns with multi-scale skeleton representations and predict the next motion with the guidance of the control signals. }
    \label{fig:overview}
\end{figure*}

\subsection{Kinematics Based Controllable Motion Synthesis}

The kinematics based methods focus on the motion of character bodies including the joints without considering the physics that cause them to move.  Four major categories of methods are reviewed as follows. 

\textit{Search-based Methods}: Early studies were based on graphs \cite{motiongraph1,motiongraph3, motiongraph2, motionfield}, where each frame of a motion database was treated as a vertex and edges represented possible transitions between two frames. A graph search can find a path to produce an expected motion. Motion matching \cite{motionmatching, learnedmotionmatching} simplified the graph search by finding transitional frames directly in animation databases and produced the state-of-art gaming animation \cite{matchingapplication2, matchingapplication3}. However, the matching criterion are often required to be devised by experienced animators for a wide range of motion scenarios.

\textit{Recurrent Neural Network based Methods}: \citet{firstRNN} constructed an encoder-decoder structure based on recurrent neural network (RNN) and directly adopted 3D body joint angles to predict character poses. \citet{improvedRNN} addressed the error accumulation issue by introducing a teacher forcing-like mechanism \cite{teacherforcing}. \citet{controllableRNN} incorporated control signals into RNNs. To alleviate the low responsiveness and slow motion transition issues caused by the inflexible RNN memory state, comprehensive data augmentation was conducted to enrich transitional patterns. However, less motion diversity was observed during the runtime, since the augmented knowledge was still limited.

\textit{Phase-based Methods}: Phase-functioned neural network \cite{PFNN} adopted a multilayer perceptron (MLP) to predict biped locomotion with an auxiliary foot contact phase, which clusters motion with similar timing to disambiguate motion predictions. The phase-based frameworks were further extended to quadruped locomotion \cite{MANN}, environmental interaction \cite{NSM}, basketball game \cite{localphase} and martial arts \cite{neurallayering}. Nevertheless, the acquisition of phase information relied on the expertise of animators and the contact information of characters, which may not be universally available. \citet{stylemodel} proposed a heuristic principal component analysis based strategy to compute the phase of a stylised motion, where the arms often exhibited special movements without contact states. However, it was still a task-specific solution.

\textit{Generative Methods}: Instead of predicting a single motion pose, modelling the conditional pose distribution and conducting sampling could avoid the averaging pose from vastly different poses \cite{recurrentVAE, motionVAE, moglow, multi-gan,  ganimator, trajevae}. \citet{motionVAE} used a variational autoencoder (VAE) to estimate the next pose distribution and draw user control-conditioned samples through reinforcement learning. Normalising flow was also introduced for this purpose, which modelled motion distribution and control signal together \cite{moglow}. Although the generative approach did not require auxiliary information, it heavily depended on balanced MoCap data distributions \cite{motionVAE}, not designed for trajectory-based control signal \cite{multi-gan} or less controlled on produced motion gait \cite{moglow}.

\subsection{Transformer in Motion Learning}
Transformers \cite{transformer} have achieved great success in a wide range of tasks such as natural language processing \cite{bert} and computer vision \cite{visiontransformer, vivit}. Compared with traditional recurrent neural networks, self-attention mechanisms perform more effectively and efficiently to address sequential patterns. Therefore, various transformer based methods were proposed for many motion related tasks such as motion prediction \cite{motionattention, posetransformer,sttransformer,mrtransformer}, action recognition \cite{st-tr, actiontransformer}, 3D pose estimation \cite{poseformer} and motion synthesis \cite{transformervae}. 
However, there are few studies utilising transformer for controllable motion synthesis. 

\subsection{Multi-scale Skeleton}
To better explore rich spatial skeleton representations of human poses, many studies introduced multi-scale skeletons by using higher order polynomials of adjacency matrices \cite{disentanglegcn}, graph convolutions \cite{motionpuzzle} or heuristic rules \cite{DMGNN, msr-gcn, compositiontext, affectivegesture}. We address the multi-scale graphs with transformers to provide multi-scale tokens with trajectories, which is the first attempt in controllable motion synthesis for more responsive motions. 

\section{Methodology}

Figure \ref{fig:overview} illustrates the proposed MCS-T architecture, which addresses the motion control problem as a regression task. The motion data is first parameterised as pose and trajectory embeddings. The pose embedding is formulated by multi-scale skeleton graphs for comprehensively exploiting the past spatio-temporal relations. It encodes the motion sequence for each skeleton scale by specialised transformer encoders for latent motion representation. The representation is then utilised by a transformer decoder queried by trajectory information, i.e., control signal, for a control-conditioned integration with past motion states. Finally, a motion prediction network predicts the character's next pose and potential future trajectory. 

\subsection{Multi-scale Skeleton Poses} \label{sec:multi-scale}

A virtual character is animated upon a skeleton of rotational joints, of which the coordinates and velocities can be defined regarding the motion.
Each pose skeleton in a motion sequence can be viewed as a graph, where the joints are vertices and the bones are edges. 
Based on such graph representation, multi-scale skeletons can be constructed for a pose by aggregating the adjacent vertices as a pooled coarse-level vertex. As illustrated in Figure \ref{fig:overview}, two scales of skeletons in a fine-to-coarse scheme are adopted in this study. This scheme aims to comprehensively characterise the spatial patterns, by which the additional coarse-level representation enables global observations of motions and improves motion dynamics especially during a motion transition.

The fine-level representation is the same as the original skeleton structure obtained from MoCap, which contains 31 joints. 
Specifically, we denote $j_{i}^p$ and $j_{i}^v$ 
as the vectors representing the coordinates local to the corresponding root transformation and velocity values of the fine-level vertices (i.e. joints) of the $i$-th frame, respectively.
For the coarse-level representation, we denote $b_{i}^p$ and $b_{i}^v$ 
as the vectors representing the coordinates and velocity values of the vertices (i.e., aggregated joints) of the $i$-th frame, respectively.

Particularly, for the motion prediction of the $i$-th frame, 
we construct an input $X_i$, which consists of two components regarding the past pose information of the two skeleton scales: $J_i$ and $B_i$. 
In detail, we denote $J_i = \{(j_{i-k}^p, j_{i-k}^v)\}$, $B_i = \{(b_{i-k}^p, b_{i-k}^v)\}$, $k=k_1,...,k_K$, as the fine and coarse sequences, respectively.
In total, $K$ frames are adopted instead of using all frames for the consideration of both efficiency purpose and model complexity.

\subsection{Multi-scale Motion Encoder} \label{sec:encoder}

A motion encoder aims to formulate the past motion patterns $X_i$ as a reference for predicting the future motion. 
Compared with the methods using auxiliary information, our encoder only depends on the past motion information available and can be generalized to all kinds of actions. However, 
trivially using all past motion information could result in issues of low responsiveness and slow motion transition. In other words, using all past information can introduce redundancy for predicting the next pose and sometimes even disturb the prediction, which may fall into the historical motion states. Thus, a transformer-based multi-scale encoder is proposed to formulate the past motion patterns in an adaptive manner.

The fine and coarse-level pose information $J_i$ and $B_i$ can be treated as matrices, where each row represents the position and velocity of a particular temporal motion frame.  
Our multi-scale encoder is based on self-attentions \cite{transformer} using the concepts of query, value and key, which can be formulated as:
\begin{equation} \label{query key value}
\begin{split}
    Q^J_i = J_i W^{Q,J}, K^J_i = J_i W^{K,J}, V^J_i = J_i W^{V,J}, \\
    Q^B_i = B_i W^{Q,B}, K^B_i = B_i W^{K,B}, V^B_i = B_i W^{V,B},
\end{split}
\end{equation}
where $W^{\cdot,\cdot}$ are projection matrices containing trainable weights with an output dimension $\gamma$, $J$ related matrices formulate the fine-level pose patterns and $B$ related matrices formulate the coarse-level pose patterns. 
Then, the temporal patterns can be computed for each level as follows: 
\begin{equation} \label{equ:attention head}
   Z^J_i = \text{softmax}(\frac{Q^J_i{K^J_i}^\intercal}{\sqrt\gamma})V^J_i, 
   Z^B_i = \text{softmax}(\frac{Q^B_i{K^B_i}^\intercal}{\sqrt\gamma})V^B_i.
\end{equation}

To this end, the temporal relations of motion frames can be formulated by observing the entire sequence based on the weights obtained using the softmax function in Eq. (\ref{equ:attention head}). In practice, multiple independent self-attentions can be adopted to increase the capability of modelling and feed-forward components are followed, which is known as a transformer encoder layer. By stacking multiple transformer encoder layers for each observation level, the final spatio-temporal patterns can be obtained. For the convenience of notations, we still use the symbols $Z^J_i$ and $Z^B_i$ to indicate the encoded sequential representations. 
By concatenating $Z^J_i$ and $Z^B_i$ in a frame-wise manner, a sequence $\{Z_i\}$ can be obtained as the encoded multi-scale past motion patterns.

\subsection{Control Signal \& Trajectory} 
\label{sec:parameter}

The trajectory of a character's movement is based on the user's control signals. 
We denote a trajectory vector:
\begin{equation}
\begin{split}
T_i = (t_{i,s-S}^p,...,t_{i,s}^p,...,t_{i,S-1}^p, t_{i,s-S}^d,...,t_{i,s}^d,...,t_{i,S-1}^d,\\ t_{i,s-S}^h,...,t_{i,s}^h,...,t_{i,S-1}^h, t_{i,s-S}^g,...,t_{i,s}^g,...,t_{i,S-1}^g),
\end{split}
\end{equation}
which represents the sampled discrete trajectory patterns for the prediction of the frame $i$. Particularly, the indices of the sampled points are specified as $s\in\{s_{-S},...,s_0,...,s_{S-1}\}$. 
In this study, we empirically adopt $S = 6$ and the sampled points are evenly distributed around the current frame to cover the trajectories 1 second before and 1 second after. 
In detail, the trajectory includes four aspects: 
\begin{itemize}
  \item $t^p_{i,s}$ represents the sampled $s$-th trajectory position in the 2D horizontal plane of the $i$-th frame. 
  \item $t^d_{i,s}$ indicates the trajectory direction in the 2D horizontal plane, which is the facing direction of the character.
  \item $t_{i,s}^h$ is a sub-vector contains the trajectory height in line with the terrain to characterise the geometry information, which are obtained from three locations regarding the sampled point including the center, left and right offset.
  \item $t^g_{i,s}$ is a one-hot encoding sub-vector regarding the action category for the sampled trajectory point. For our locomotion settings, we have five action categories including standing, walking, jogging, jumping and crouching.
\end{itemize}

\subsection{Control Signal-aware Decoder} 
\label{sec:decoder}

Based on the past motion embeddings from the multi-scale motion encoder, a control signal-aware decoder is proposed to formulate a latent embedding for motion prediction. The trajectory information is involved by the decoder to attend to the past encoded motion patterns through a control signal-aware attention mechanism. This allows the decoded patterns being relevant to the user's control signals. In detail, we adopt the trajectory $T_i$ as a query to the past motions: 
\begin{equation} \label{equ:decoder}
\begin{split}
q^D_i = T_i W^{Q,D},K^D_i = Z_i W^{K,D},V^D_i = Z_i W^{V,D},
\end{split}
\end{equation}
where $W^{\cdot,\cdot}$ are projection matrices containing trainable weights with an output dimension $\gamma$. Hereafter, the past motion information with user control can be summarised into a vector as follows:
\begin{equation} \label{equ:decoder attention head}
   z^D_i = \text{softmax}(\frac{q^D_i{K^D_i}^\intercal}{\sqrt\gamma})V^D_i.
\end{equation}
Particularly, we call the attention in Eq. (\ref{equ:decoder}-\ref{equ:decoder attention head}) as a control signal-aware attention and multi-heads of it are adopted with feed-forward networks to characterise the motions from multiple aspects. 
For the simplification of notations, we continue to use $z_i^D$ to denote this multi-head output.

\subsection{Motion Prediction Network}

To predict and synthesize the motion of the $i$-th frame, which we denote as $Y_i$, an additional motion prediction network (MPN) component is introduced.
${Y}_{i}$ contains pose $\{ (j_{i}^p, j_{i}^v, j_{i}^r) \}$, trajectory $T_{i+1}$ and contact information $C_{i}$. Particularly, $j_{i}^r$ represents local joint rotation additional to position and velocity. The prediction $\hat{T}_{i+1}$ of $T_{i+1}$ is only for the trajectory after the $i$-th frame, where the sampled trajectory points before the current frame already exist. $C_{i}$ is a vector, which indicates the labels of foot contact for each heel and toe joint of the two feet. It can be used to perform Inverse Kinematics (IK) post-processing to better fit the character with terrain geometry. 

Our MPN is based on
feed-forward layers with Exponential Linear Unit (ELU) activation function \cite{elu}. 
In detail, we have an estimation $\hat{Y}_i$ of $Y_i$: 
\begin{equation} \label{equ:mpn}
   \hat{Y}_i = \text{MPN}(z_i^D, T_i),
\end{equation}
where the decoded output and motion trajectory are considered as the input. Note that the trajectory information is also used for MPN besides the decoder, which helps the control signals to be fully formulated for providing highly responsive motion synthesis.

\subsection{MCS-T Training and Runtime Inference}

By defining the computations of the proposed MCS-T as a function $\mathcal{F}$ with trainable parameters $\boldsymbol{\Theta}$, where $\hat{Y}_i = \mathcal{F}(X_i, T_i)$. A mean squre error (MSE) loss with $\ell_1$ regularization is adopted to optimize $\boldsymbol{\Theta}$. In detail, we solve the following optimization problem during the training:
\begin{equation} \label{cost}
    \argmin_{\boldsymbol{\Theta}} \parallel {Y}_{i} - \boldsymbol{\mathcal{F}}({X}_{i}, {T}_{i};\boldsymbol{\Theta}) \parallel^2_2 + \lambda |\mathbf{\boldsymbol{\Theta}|},
\end{equation}
where $\lambda$ is a hyper-parameter controlling the scale of the regularization. 

In terms of the runtime inference, a trajectory blending scheme is adopted for post-processing. 
In detail, 
the trajectory positions $\hat{t}_{i+1,s}^p$ and directions $\hat{t}_{i+1,s}^d$,  $s=s_0,...,s_{S-1}$, after the $i$-th frame are further blended with the user control signal for the $(i+1)$-th frame's motion prediction:
\begin{equation} \label{equ:blending}
\begin{split}
    t^p_{i+1,s} = (1-\tau_s^p) \bar{t}^p_{i+1,s} + \tau_s^d \hat{t}^p_{i+1,s}, \\
    t^d_{i+1,s} = (1-\tau_s^d) \bar{t}^d_{i+1,s} + \tau_s^d \hat{t}^d_{i+1,s},
\end{split}
\end{equation}
where $\bar{t}_{i+1,s}$ is the trajectory computed by the user's control signal, $\tau_s^p$ and $\tau_s^d$ are hyper-parameters to control the blending level. That is, 
the user control signal is blended with higher weights in near trajectory for more responsive motion, and with lower weights in far trajectory in pursuit of smoother transition. 
In terms of $t^p_{i+1,s}$ and $t^d_{i+1,s}$, $s=s_{-S},...,s_{-1}$, they are in line with the actual existing trajectory. 
Additionally, the trajectory height $t_{i+1,s}^h$ can be derived based on $t^p_{i+1,s}$ within the virtual scene, and the action category $t_{i+1,s}^g$ is set directly by the user.

\section{Experimental Results and Discussions}

\subsection{Dataset}

We evaluate our proposed method on a public dataset \cite{PFNN} for a fair comparison with the state-of-the-art methods. The dataset consists of biped locomotion data of various gaits, terrains, facing directions and speeds, which helps evaluate the quality of the common character motion controller in terms of responsiveness and motion transition. 
A biped character with 31 joints and MoCap techniques were adopted to collect these data. In total, we obtained around 4 million samples for training. 

\subsection{Implementation Details}

In total, $K=5$ past frames with indices $k_1= 1$, $k_2= 10$, $k_3= 20$, $k_4= 30$ and $k_5= 40$ were selected as input to predict the motion of the $i$-th frame. Note that this setting was found to provide the best quality of prediction (see Section \ref{sec:discussion}). 
Two independent transformer-encoders were used for the fine-level and coarse-level motion sequences, respectively. Each of them consisted of three transformer-encoder layers using six self-attention heads of a dimension 186 and the the feed-forward layers were of a dimension $1024$. A dropout rate of $0.1$ was applied to the encoders. The transformer decoder was using the same configurations as the encoder. The motion prediction network was modelled as a three-layer MLP with a hidden dimension $512$ and a dropout rate $0.3$. $\tau^p_s = (s/S)^{0.5}$ and $\tau^d_s = (s/S)^{2}$ were defined for Eq(\ref{equ:blending}) empirically. (see Appendix for details)

During the training, the input and output were firstly normalised by their mean and standard deviation. Additionally, the input features related to the joints were all scaled by $0.1$, which helped produce dynamic motions in certain scenarios to enlarge the proportion of the trajectory related inputs.   
In terms of the loss function, $\lambda$ for $\ell_1$ regularization was set to 0.01. 
The model was implemented by PyTorch \cite{pytorch} and trained with an Adam optimisier \cite{adam}.  The learning rate was set to  $10^{-4}$ and the batch size was 32. In total, MCS-T was trained with 20 epochs, which took around 50 hours on an Nvidia GTX 1080Ti GPU.

\subsection{Comparisons with State-of-the-art Methods} \label{sec:evaluation}

Qualitative and quantitative evaluations of MCS-T were conducted against a number of baseline methods, in terms of motion quality, especially from the aspects of responsiveness and motion transition. The baseline methods include MLP with a single past pose, MLP with multiple past poses, RNN \cite{controllableRNN} and PFNN \cite{PFNN} methods. Overall, 
we show that our MCS-T was able to produce motions in line with the-state-of-the-art results with a task-agnostic design and to alleviate the fundamental issues of the baseline methods. More results are available in the supplementary material. 

\begin{figure*}[t]
    \centering
    \includegraphics[width=\linewidth]{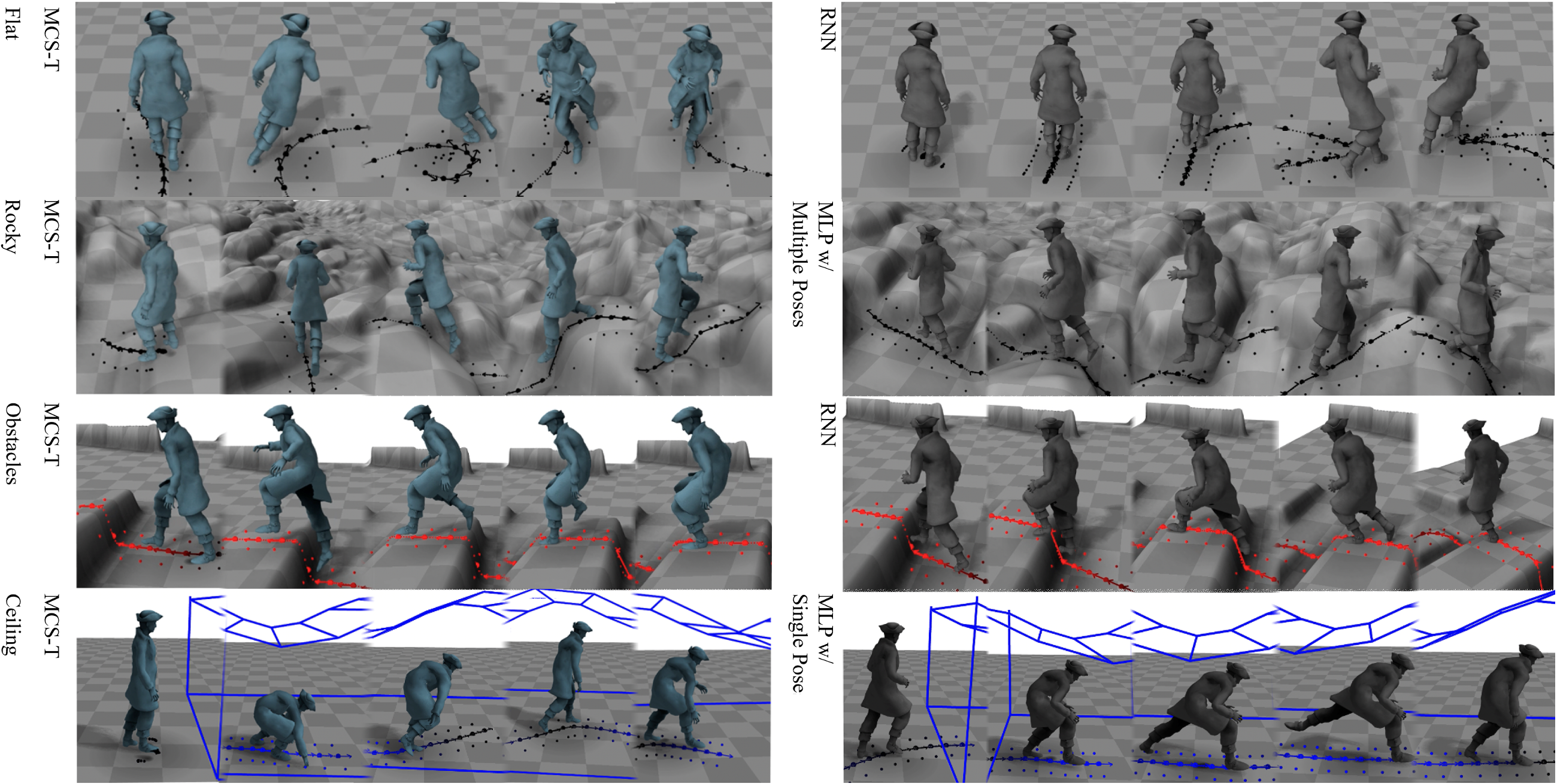}
    \caption{Qualitative results of our MCS-T and other baselines under four scenarios: flat, rocky, obstacles and ceiling. The left side shows the motions synthesized by MCS-T and the right side provides examples that demonstrate the baseline limitations. }
    \label{fig:comparison}
\end{figure*}

\textit{MLP with single past pose}: We trained an MLP to synthesize motion using a single past pose with trajectory information. The experimental results show that the overall motion produced was quite stiff especially when changing the direction and could have weird artifacts such as floating as shown in the ceiling scenario in Figure \ref{fig:comparison}. As expected, motion prediction from vague control signal can be difficult without auxiliary information and various possible predictions can exist, which leads to an average pose.

\textit{MLP with multiple past poses}: Similar to the first baseline, except that additional pose information from multiple past frames was considered for an MLP. The results show that the generated motion was improved, as the past frames provided the auxiliary information implicitly. 
Nonetheless, the synthesized motion suffers from the slow motion transition issue. This problem became obvious when the character was traversing through rocky terrain as shown in Figure \ref{fig:comparison}. While it was able to adapt the character motion correctly to the new geometry, the motion was performed as smooth as the regular locomotion on a flat terrain. The reason could be that using several past poses in a simple manner is limited to the large redundant variations in the past.

\textit{RNN}: An LSTM architecture \cite{controllableRNN} was adopted for this dataset without augmentation. The past memory enables LSTM to predict motion of higher quality. Nevertheless, it still suffered from the slow motion transition issue. As shown in Figure \ref{fig:comparison}, the character could be floating when transiting between motion and was unable to jump over the obstacles timely and obviously. The reason is that the hidden memory prevented the RNN model from quickly reaching transitional states of a jumping motion. 

\textit{Phase-functioned neural network}: Rather than relying on past poses to constrain motion prediction, PFNN \cite{PFNN} utilised the foot contact phase for motion disambiguation. The qualitative results of our MCS-T were very closely to those of PFNN in a wide range of scenarios (see supplementary materials). Our method does not require the task-specific auxiliary information, which only relies on the past motion data generally available. 

Moreover, to quantitatively evaluate whether the produced motion is responsive to control signals and transits to different motions timely, the average joint angle update per second as a metrics for motion dynamics was compared. A higher joint angle update represents more dynamic motion produced and faster transition between frames. The results are listed in Table \ref{tab:angleupdate}, which indicate that MCS-T is able to produce much more agile motion than the task-agnostic RNN, while being comparable to the task-specific PFNN \cite{PFNN} method. 

\begin{table*}[t]
    \centering
    \resizebox{\textwidth}{!}{
    \begin{tabular}{p{4.6cm} |c| p{0.7cm}<{\centering} p{0.7cm}<{\centering} p{0.7cm}<{\centering} p{0.7cm}<{\centering} p{0.7cm}<{\centering} p{0.7cm}<{\centering} p{0.7cm}<{\centering} p{0.7cm}<{\centering} p{0.7cm}<{\centering} p{0.7cm}<{\centering} p{0.7cm}<{\centering} p{0.7cm}<{\centering} |p{0.7cm}<{\centering} p{0.7cm}<{\centering} p{0.7cm}<{\centering}}
        & & \multicolumn{3}{c}{Flat} & \multicolumn{3}{c}{Rocky} & \multicolumn{3}{c}{Obstacles} & \multicolumn{3}{c|}{Ceiling} &  \multicolumn{3}{c}{Average} \\
        Method & Phase & Full & Arm & Leg & Full & Arm & Leg & Full & Arm & Leg & Full & Arm & Leg & Full & Arm & Leg  \\
    \hline
        PFNN &\checkmark & \underline{106.5} & 100.6 & \underline{135.9} & \textbf{128.7} & \underline{145.0} & \textbf{156.0} & \underline{109.1} & \underline{110.9} & \underline{143.9} & \underline{139.9} & 130.9 & \textbf{187.0} &  \underline{121.1} & \underline{121.9} & \underline{155.7} \\
        MLP w/ Single Pose &\XSolidBrush  & 71.5 & 65.4 & 90.2 & 86.5 & 90.3 & 110.5 & 78.7 & 71.2 & 108.9 & 109.7 & 103.7 & 142.0 & 86.6 & 82.7 & 112.9 \\
        MLP w/ Multiple Poses &\XSolidBrush  & 94.0 & 88.1 & 122.7 & 95.2 & 91.3 & 131.0 & 85.7 & 76.3 & 122.6 & 115.1 & 100.8 & 161.8 & 97.5 & 89.1 & 134.5 \\
        RNN &\XSolidBrush  & 83.3 & 78.4 & 107.1 & 83.2 & 76.4 & 115.5 & 85.5 & 80.4 & 122.3 & 123.7 & 107.9 & 174.2 & 93.9 & 85.8 & 129.8 \\
    \hline\hline
        MCS-T (ours)&\XSolidBrush  & \textbf{110.9} & \textbf{107.5} & \textbf{142.8} & \underline{126.7} & \textbf{149.0} & \underline{151.6} & \textbf{116.1} & \textbf{121.4} & \textbf{150.7} & \textbf{140.0}  & \underline{137.0} & \underline{184.6} & \textbf{123.4} & \textbf{128.7} & \textbf{157.4}\\
        
        + Middle scale&\XSolidBrush  & 105.5 & \underline{101.6} & 135.4 & 109.2 & 117.2 & 140.8 & 104.6 & 108.6 & 140.1 & 122.0 & 111.5 & 164.5 & 110.3 & 109.7 & 145.2 \\
        - Multi-scale skeleton &\XSolidBrush  & 96.3 & 87.6 & 127.1 & 112.6 & 123.8 & 143.4 & 91.9 & 89.4 & 127.2 & 124.1 & 114.1 & 167.1 & 106.2 & 103.7 & 141.2 \\
        - Control signal-aware attention &\XSolidBrush  & 94.6 & 87.7 & 125.2 & 105.6 & 109.1 & 140.2 & 90.7 & 85.2 & 129.3 & 137.5 & \textbf{145.8} & 173.7 & 107.1 & 107.0 & 142.1\\
    \end{tabular}
    }
    \caption{Quantitative comparison in terms of the average joint angle update per second (degree/s) $\uparrow$ for different methods including MCS-T under four motion scenarios: Flat, Rocky, Obstacles, and Ceiling. The angle updates are further divided into full body with all joints, arm and leg joints. The highest value is in bold and the second highest value is underlined.}
    \label{tab:angleupdate}
\end{table*}

\subsection{Ablation Study} 
\label{sec:ablation}

An ablation study was conducted to demonstrate the effectiveness of the multi-scale skeleton representation and the control signal-aware mechanism in our encoder and decoder, respectively. The quantitative evaluation is listed in Table \ref{tab:angleupdate}. 

\textit{Multi-scale skeletons with an extra middle scale}: In addition to the two skeleton scales, we experimented with one extra scale called as a middle scale. It aggregated the joints into a level between the two existing levels. However, the three-scale scheme did not contribute to the overall performance and produced stiff motions especially under scenarios with quick and frequent transitions such as obstacles and ceiling scene. The potential reason could be that the increased model complexity deteriorates the capability of motion prediction and produces sub-optimal solution.

\textit{Multi-scale skeletons}: Without multi-scale skeletons, the motion dynamics dropped significantly, especially in the obstacles scene. Jumping motion became less responsive and sometimes the dynamics were too weak to observe. Thus, incorporating coarse-level skeletons helped exploit the motion patterns during a transition from a global perspective. 

\textit{Control signal-aware decoder}: Besides the motion prediction network, our decoder is driven by the control signals as well, which are adopted as the queries of the decoding attentions. Alternatively, by simply using a conventional self-attention mechanism to construct this decoder, it led to less motion dynamics. The most obvious case is in the ceiling scenario, where the motion appeared to be jittery and unstable during the transition between the walking and the crouching in the ceiling scenario. 

\subsection{Multi-scale and Control Signal-aware Motion Attentions} \label{sec:discussion}

\begin{figure}[t]
    \centering
    \includegraphics[width=\linewidth]{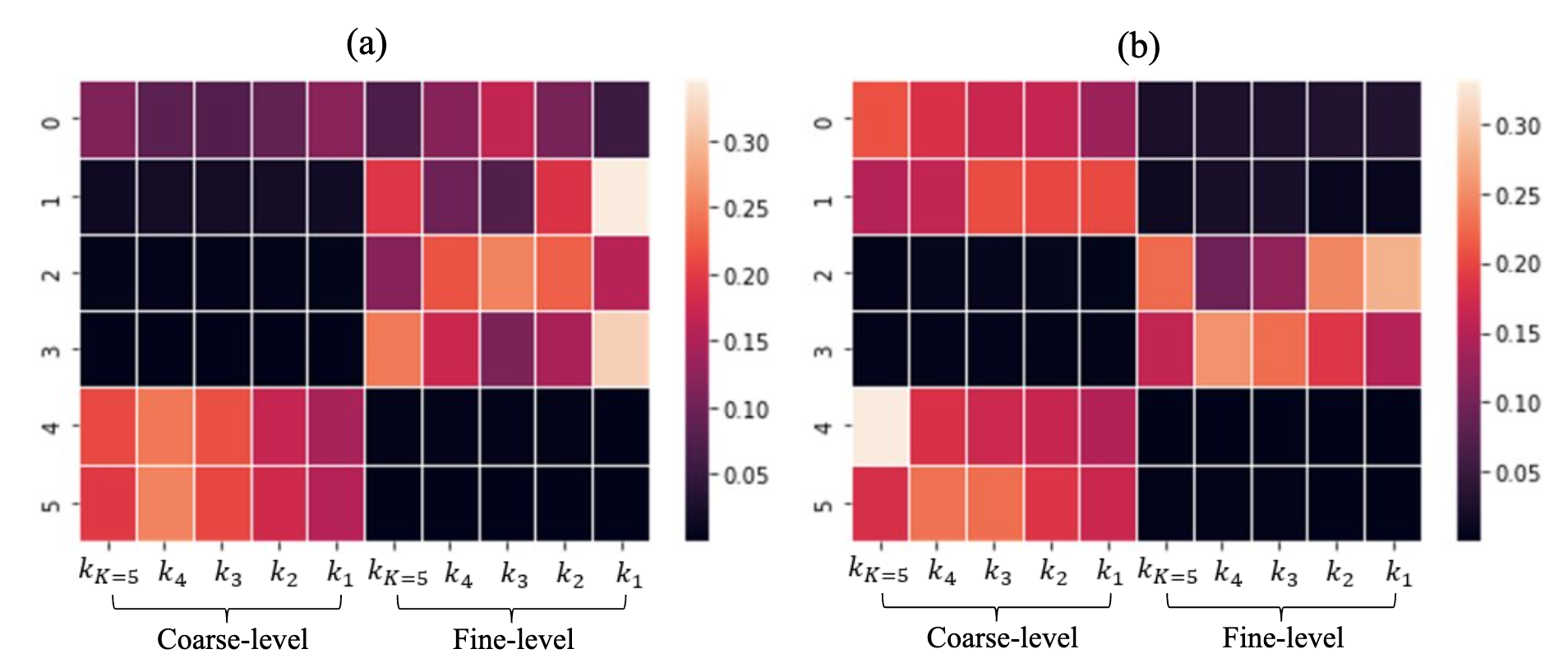}
    \caption{Visualization of the attention map of MCS-T, where the first layer of the decoder is shown, including (a) transitional and (b) non-transitional scenarios. The x-axis represents the past motion indices and the y-axis indicates the 6 attention heads.}
    \label{fig:attention}
\end{figure}

Our experiments show that MCS-T is able to synthesize motions with the highest quality and alleviate the slow transition issue. This lies in the attention mechanisms of MCS-T, which adaptively addresses the sequential motion context. 
The attention map of the decoder's first layer is visualised in Figure \ref{fig:attention} to show how MCS-T performs attentions for different cases. 
Figure \ref{fig:attention} (a) is for a frame of motion transition from a jumping state to a jogging state. Most attention heads focused on the fine-level skeletons, especially in more recent frames, as the further past frames were not very relevant during this motion transition. Additionally, two attention heads paid even attentions to the coarse-scale motion, which learned global motion patterns for faster motion transition. 
Figure \ref{fig:attention} (b) is for a non-transitional case where the character remains the jogging state, the attentions are evenly distributed on all positions of the past poses, especially with more attention heads focusing on the coarse-level. The reason could be that the coarse motion sequence provides sufficient spatio-temporal patterns for predicting this kind of motions with strong recurring patterns. 

\subsection{Limitations \& Future Work}

\begin{figure}[t]
    \centering
    \includegraphics[width=\linewidth]{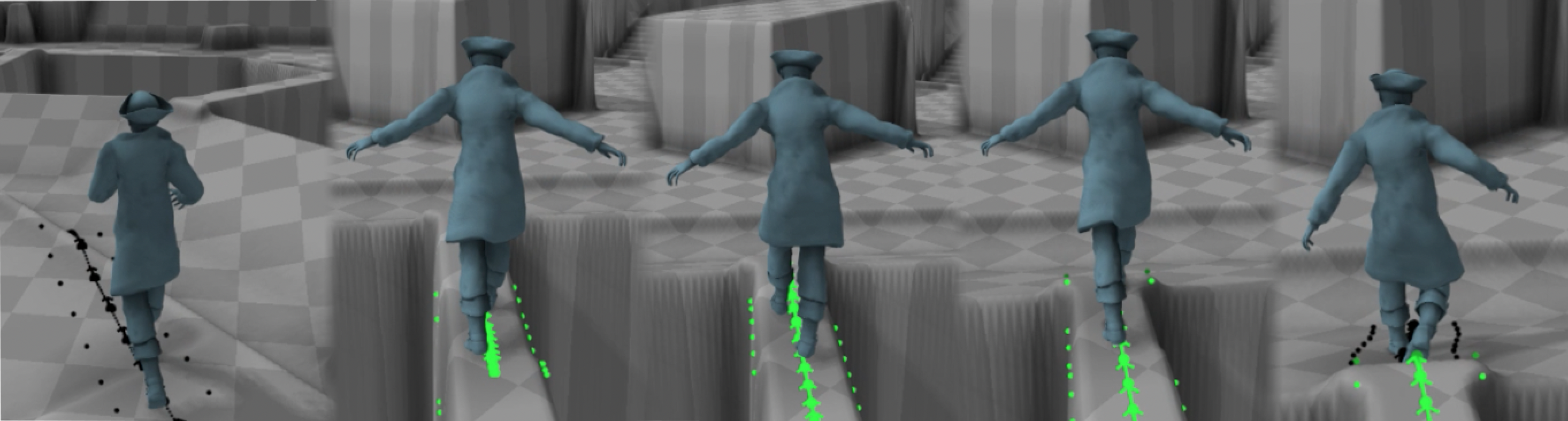}
    \caption{Illustration of a limitation of MCS-T, where the hand balancing motion is not well synthesized when the character is walking on a beam.}
    \label{fig:balancing}
\end{figure}

There are two major limitations of our proposed MCS-T. 
First, our MCS-T method may not always synthesize the beam walking motion well. For example, as shown in Figure \ref{fig:balancing}, informed by the special terrain geometry, the character performed a hand balancing motion. However, MCS-T did not always launch this motion. It could be due to the small percentage of beam walking motion in the training data (\~2\%) and imbalance learning strategies should be considered. 
Second, since MCS-T exploits the past motion history, the error accumulation could happen with a very low chance. The character motion could get stuck in weird poses for a very short period but can escape from it by providing new control signals. Robust noise-based learning could be conducted for alleviating such error accumulation.
In our future work, besides addressing these limitations, we will investigate an adaptive strategy for selecting past frames, such as exploring network architecture search (NAS) \cite{autopytorch} and token evaluation strategies.

\section{Conclusion}

In this paper, we present MCS-T as a transformer-based task-agnostic character motion control method. With multi-scale graph representation, it aims to produce responsive and dynamic motions without explicitly using auxiliary information. 
Specifically, MCS-T involves an encoder-decoder design, where the encoder formulates the spaio-temporal motion patterns of past poses from multi-scale perspectives and the decoder takes a control signal into account for predicting the next pose. 
Our experiments on a public dataset have demonstrated that MCS-T can produce results comparable to those of the state-of-the-art methods which explicitly using auxiliary information. We also investigate the limitations of our method for future improvement.

\section*{Acknowledgments}
This study was partially supported by Australian Research Council (ARC) grant \#DP210102674.

\bibliography{references.bib}

\clearpage
\begin{appendices}
\section{Average Joint Angle Update}
We define the computations of average joint angle update for quantitative evaluation in this section. Let $j^r_i$ and $j^r_{i+1}$ be the local joint rotation in unit quaternion representation at frame $i$ and $i+1$. The average joint angle update can be obtained as:

\begin{equation}
    \text{update} = \frac{1}{|N||J|} \sum_{i=1}^N \sum_{j=1}^J 2\arccos(|j^r_i \cdot (j^r_{i+1})^{-1}|).
\end{equation}
The equation above gives the average joint angle update as radian per frame. It is further converted to degree per second as presented in the quantitative evaluation. 

\section{Statistical Test}
To further validate that MCS-T is showing comparable quantitative performance to PFNN \cite{PFNN}, a paired sample t-test was conducted. The results are summarised in Table  \ref{tab:t-test}. With an alpha value 0.1, we can say the improvement is significant in mean for flat and obstacles scenes by our method. In addition, we can state that there is no significant difference in mean for rocky and ceiling scene between the two methods. 

\begin{table}[h]
    \centering
    \begin{tabular}{p{2.8cm} | p{1.5cm}<{\centering}  p{1.5cm}<{\centering}}
    Statistics & PFNN & MCS-T \\
    \hline\hline
    Flat mean & 105.5 & 109.8 \\
    Flat $p$ value & \multicolumn{2}{c}{0.036}  \\
    \hline
    Rocky mean & 125.1 & 122.5 \\
    Rocky $p$ value & \multicolumn{2}{c}{0.164} \\
    \hline
    Obstacles mean & 108.5 & 115.5 \\
    Obstacles $p$ value & \multicolumn{2}{c}{0.0001} \\
    \hline
    Ceiling mean & 143.4 & 143.9 \\
    Ceiling $p$ value & \multicolumn{2}{c}{0.799} \\
    \end{tabular}
    \caption{The paired sample t-test on PFNN and MCS-T under different animation scenes.}
    \label{tab:t-test}
\end{table}

\section{Hyper-Parameters}
Different hyper-parameter settings regarding our model has been evaluated to demonstrate the current optimal choice. The evaluation of models under different settings are summarized in the Table \ref{tab:hyperparameter}. 

\begin{table}[h]
    \centering
    \begin{tabular}{p{2.5cm} | 
    p{0.7cm}<{\centering} p{0.7cm}<{\centering} p{0.7cm}<{\centering} p{0.7cm}<{\centering} |p{0.7cm}<{\centering}}
        Method & Fl. & Ro. &  Ob. & Ce. & Aver. \\
    \hline\hline
        MCS-T (ours)   & \textbf{110.9} & \textbf{126.7} & \textbf{116.1} & 140.0 & \textbf{123.4}\\
        \hline
        w/ MAE loss & 110.6 & 116.5 & 86.1 & 98.6 & 103.0\\
        w/ CE loss & 105.7 & 118.5 & 109.0 & \textbf{140.2} & 118.4\\
        w/ $\tau^p_s = (s/S)^{2}$, & \multirow{2}{*}{108.2} & \multirow{2}{*}{112.1} & \multirow{2}{*}{108.4} & \multirow{2}{*}{138.8} & \multirow{2}{*}{116.9}\\
        $\tau^d_s = (s/S)^{5}$ & & & & &\\
    \end{tabular}
    \caption{Average joint angle update per second (degree/s) $\uparrow$ for our model with different settings in all four scenarios}
    \label{tab:hyperparameter}
\end{table}

First, we change the MSE in the loss function - Equation \ref{cost} with an MAE for possible improvement of robustness. However, it results in less dynamic motion in all scenarios. We suppose that the worse performance is due to the scale-invariant nature of MAE gradient. In addition, MAE focuses more on outliers, whilst human pose attributes are generally within a particular range. Thus, for robust learning, suitable mechanisms could be further studied for the small error accumulation and noise rather than outliers during motion synthesis. 

Next, the cross entropy (CE) loss was specifically adopted for the foot contact label while MSE for other outputs. Although CE loss could be more suitable for a binary label, this setting produced less dynamic motion and even floating motion. It is possibly because this different CE loss term require more carefully tuned regarding its importance. 

Finally, we evaluate different hyper-parameter setting of Equation \ref{equ:blending}. With the new setting, the trajectory blending is more towards the gamepad control signal. As a result, the character motion is more responsive to the gamepad control such as directional instruction reflected by qualitative evaluation. However, it comes with the costs of less dynamic motion during transition which leads to a decrease of average joint angle update. This trade-off between relevant hyper-parameters requires animators to involve according to their specific demands.

\end{appendices}
\end{document}